\pdfoutput=1
\documentclass[letterpaper, 10 pt, conference]{ieeeconf}  

\IEEEoverridecommandlockouts                              

\title{\LARGE \bf
Is Discretization Fusion All You Need for Collaborative Perception?}

\author{ 
    Kang Yang$^{1,3}$\hspace{10pt} 
    Tianci Bu$^{2}$\hspace{10pt}
    Lantao Li$^{3}$\hspace{10pt}
    Chunxu Li$^{1}$\hspace{10pt}
    Yongcai Wang$^{*1}$\hspace{10pt}
    Deying Li$^{1}$
\thanks{$^{1}$ School of Information Renmin University of China, Bei Jing, China, 100872. \{yangkang1205, ycw, deyingli\}@ruc.edu.cn / lichunxu@wti.ac.cn}
\thanks{$^{2}$ National University of Defense Technology, Hu Nan, China, 410073. btc010001@gmail.com }
\thanks{$^{3}$ Sony Research and Development Center China, Beijing, China lantao.li@sony.com}
\thanks{$^{*}$ Corresponding author.}
}

\usepackage{multicol}
\usepackage{multirow}
\usepackage[utf8]{inputenc} 
\usepackage[T1]{fontenc}    
\usepackage{hyperref}       
\usepackage{url}            
\usepackage{booktabs}       
\usepackage{amsfonts}       
\usepackage{nicefrac}       
\usepackage{microtype}      
\usepackage{xcolor}         
\usepackage{graphicx}
\usepackage{amsmath}
\usepackage{colortbl}
\renewcommand{\arraystretch}{1.0} 
\usepackage{algorithm}
\usepackage{algpseudocode}
\usepackage{pifont}
\usepackage{caption}
\usepackage{wrapfig}  
\usepackage{footnote} 
\usepackage{float}
\usepackage{adjustbox}  
\usepackage{listings}

\usepackage{subcaption}
\usepackage{hyperref}
\usepackage{rotating}

\begin{document}

\maketitle


\begin{abstract}
Collaborative perception in multi-agent system enhances overall perceptual capabilities by facilitating the exchange of complementary information among agents. 
Current mainstream collaborative perception methods rely on discretized feature maps to conduct fusion, which however, lacks flexibility in extracting and transmitting the informative features and can hardly focus on the informative features during fusion. To address these problems, this paper proposes a novel Anchor-Centric paradigm for Collaborative Object detection (ACCO). It avoids grid precision issues and allows more flexible and efficient anchor-centric communication and fusion. ACCO is composed by three main components: (1) Anchor featuring block (AFB) that targets to generate anchor proposals and projects prepared anchor queries to image features. (2) Anchor confidence generator (ACG) is designed to minimize communication by selecting only the features in the confident anchors to transmit. (3) A local-global fusion module, in which local fusion is anchor alignment-based fusion (LAAF) and global fusion is conducted by spatial-aware cross-attention (SACA). LAAF and SACA run in multi-layers, so agents conduct anchor-centric fusion iteratively to adjust the anchor proposals. Comprehensive experiments are conducted to evaluate ACCO on OPV2V and Dair-V2X datasets, which demonstrate ACCO's superiority in reducing the communication volume, and in improving the perception range and detection performances. Code can be found at: \href{https://github.com/sidiangongyuan/ACCO}{https://github.com/sidiangongyuan/ACCO}.
\end{abstract}

\section{Introduction}

3D object detection through the collaboration of multiple agents is a crucial problem for accurate and reliable autonomous driving. It has attracted significant attention in recent years. Multi-agents offer varied viewpoints to efficiently overcome the inherent limitations of single-agent perception, such as occlusion and long range issues. Currently, numerous effective collaboration methods \cite{where2comm,CoBEVT,v2vnet,v2x-vit} and high-quality datasets \cite{OPV2V,dair-v2x,v2v4real,v2x-vit} have been introduced. 




However, as we know, current mainstream collaborative perception methods, whether Lidar-based \cite{where2comm,OPV2V,v2vnet} or Camera-based \cite{CoBEVT,CoCa3D}, conduct fusion by discretized features received from neighboring agents, which is called the Discretization Fusion (DF) paradigm. 
Although DF is intuitively reasonable and straightforward, it faces two inevitable problems. First, it needs to trade off among the precision of the grids, the encoded range in the map, and the computation and communication cost. Fusion fine-grained and large scale feature maps are costly, therefore existing methods \cite{where2comm,CoBEVT,v2vnet,v2x-vit} adopt downsampling for feature maps. Secondly, the feature map contains a large amount of redundant background information, which not only increases the communication volume but also introduces noises into the fusion process. These challenges hinder the agents' ability to effectively integrate features with the collaborative agents. This raises a pivotal question:
\textit{Is discretization fusion all we need for collaborative perception?}

We propose an Anchor-Centric paradigm for Collaborative Object detection (ACCO), which is a novel paradigm that employs a DETR \cite{detr,Detr3D} structure to detect and fuse information all through anchor queries. 
The core concept is to initially randomly generate, and iteratively refine a set of anchor queries and the corresponding anchor features at each agent. In collaboration, we use the spatial position information of the anchor queries to guide the fusion of the anchor features. Each agent only transmits high-quality anchors, enabling efficient and flexible communication and fusion.

Specially, ACCO consists three core components. 
(1) the Anchor Featuring Block (AFB) projects each 3D anchor query onto the agent's surround-view or front-view image plane to extract the image feature for each anchor query.
(2) the Anchor Confidence Generator (ACG) block evaluates each anchor query by a confidence score, based on which, a fixed number of high-quality anchor queries are selected for communication. 
(3) At the ego agent, the received anchor queries from neighbors are transformed into the same coordinate system of the ego agent. These valuable anchor queries are finally fused by multiple-layer fusion module, which contains anchor-centric local fusion (LAAF) and spatial-awareness cross-attention (SACA) based global fusion. 

\begin{figure}[!tbp]
    \includegraphics[width=\linewidth]{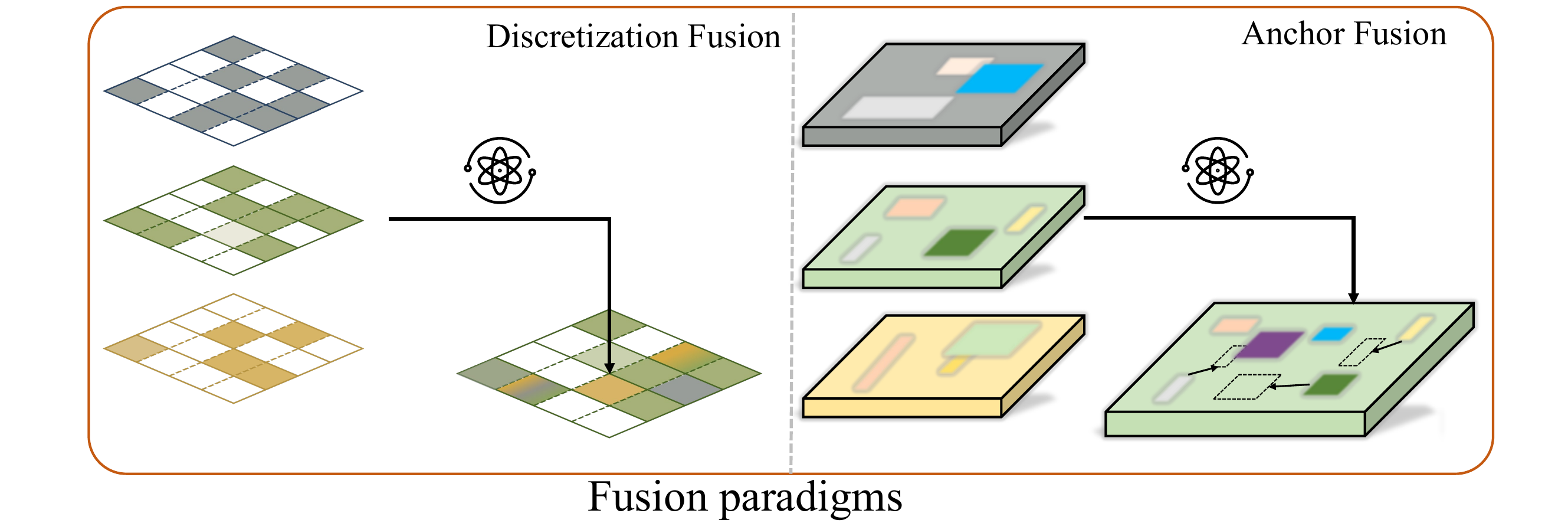}
    \caption {Fusion process in DF v.s. ACCO.  ACCO is more flexible, accurate, and efficient.}
    \label{fig:intro}
\end{figure}

Fig. \ref{fig:intro} illustrates the differences between conventional fusion and ours. The anchor-centric fusion breaks the border of grids, which is more flexible, efficient, accurate, and is also more explainable.  
To assess the performances of ACCO, we leverage two widely used datasets for evaluation: Dair-v2x \cite{dair-v2x} (real-world) and OPV2V \cite{OPV2V} (simulation scenarios).

\section{Related work}
\textbf{Collaborative Perception.} 
Collaborative perception is a highly promising application in multi-agent systems that aims to integrate data from multiple agents to improve the precision of 3D object detection. Exiting data fusion strategies can be categorized into three types: early fusion, intermediate fusion and late fusion. Early fusion strategy \cite{Cooper1,Cooper2}, employs raw sensor data from collaborative agents and achieves exemplary performance. While this strategy is simple and straightforward, it requires a significant amount of communication bandwidth. Late fusion employs prediction fusion at the network output, making it more bandwidth-efficient and simpler than early and intermediate collaboration. However, its outputs can be noisy and incomplete, often resulting in the worst perception performance. To solve the drawbacks of the previous two methods, intermediate fusion stands as the predominant strategy. V2VNet \cite{v2vnet} uses GNNs to first compensate for time delay. V2X-VIT \cite{v2x-vit} utilizes an attention mechanism covers V2V and V2I simultaneously. Where2comm \cite{where2comm} addresses where fusion needs to occur to reduce communication bandwidth. DiscoNet \cite{disconet} employs knowledge distillation to leverage the benefits of both early and intermediate collaboration. CoBEVT \cite{CoBEVT} presents the first generic, multi-camera-based collaborative perception framework for cooperative BEV (Bird's Eye View) semantic segmentation. HEAL \cite{HEAL} answers the question of how to accommodate continually emerging new heterogeneous agent types into collaborative perception. Previous works mainly focus on Discretization Fusion (i.e. feature map fusion), while we propose a completely different fusion paradigm, achieving collaborative perception from another perspective.


\textbf{Camera-Based BEV perception.} Camera-Based perception methods are increasingly popular since camera sensors are significantly more cost-effective compared to LiDAR systems \cite{bevwork1,bevwork2}. In recent years, the BEV paradigm has gained prominence, with many camera-based 3D perception methods yielding promising results. Predominantly, Camera-based BEV methodologies fall into two categories: Bottom-up (forward projection) and top-down (backward projection) paradigms. Bottom-up, exemplified by works like LSS \cite{LSS,BEVDet,LSS-1,LSS-2,LSS-3}, first estimate the depth distribution and then project the 2D image features along the distribution ray to obtain the 3D voxel features, which are then collapsed to BEV features. The performance of the system is intimately related to the accuracy of depth estimation. As a result, some studies like BEVDepth \cite{BEVDepth,CaDNN} have incorporated depth supervision signals from Lidar to guide depth prediction. Another approach, the top-down paradigms (backward projection), such as BEVFormer \cite{BEVFormer,Detr3D,petr,sparse4D}, primarily utilises the concept of DETR \cite{detr}, using a transformer-based framework. It designs a set of prior queries in BEV space and then using an inverse projection, the queries are projected onto the image feature plane to sample features. This paradigm can avoid depth prediction, but it requires greater computational resources and is difficult to converge. In this paper, we choose the top-down approach as our basic backbone. Firstly, it avoids the need for precise depth prediction. Secondly, this method enables more flexible and efficient anchor-centric fusion.

\section{Method}

\begin{figure*}[htbp]
    \includegraphics[width=\textwidth]{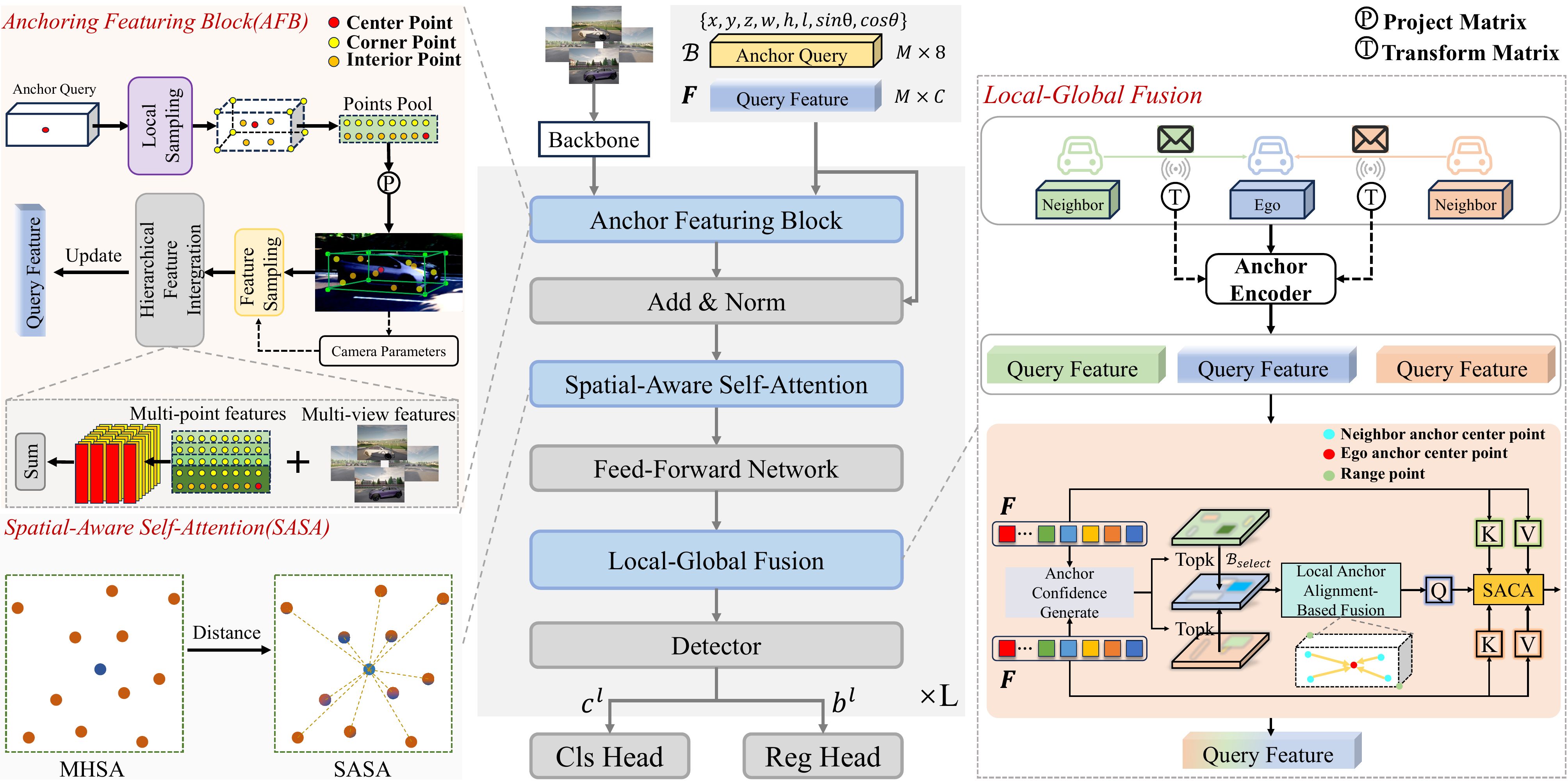}
\caption{Framework. The encoder layer of ACCO contains anchor queries, anchor featuring block, spatial-aware self-attention, and local-global fusion. Anchor queries are initialized as a sparse set of proposals in the BEV space. The spatial-aware self-attention encodes the queries with spatial distance. Local-global fusion is a critical component comprising several key elements: the anchor encoder, anchor confidence generator, local anchor alignment-based fusion, and spatial-aware cross-attention. The decoder repeats $L$ times to produce final predictions.}
  \label{Framework}
\end{figure*}
This work exploits the top-down BEV perception paradigms to present the first anchor-centric fusion framework. As illustrated in Fig. \ref{Framework}, the input consists of features extracted from multi-view images using the backbone, along with a predefined set of anchor queries and their corresponding query features. For specific details about the query, please refer to sections \ref{expsetting} and \ref{query defined}. The AFB module updates the anchor features by allowing interaction between the image features and the anchor queries, achieving the goal of anchor featuring. The SASA part is an attention mechanism specifically designed for anchor queries, taking spatial distance into account. Immediately following this, Local-global fusion, a pivotal component of our paper, is tailored for the efficient and flexible fusion of multi-agent systems. Finally, the detector decodes the classification and regression results, refining the anchor queries progressively through each layer. The entire structure comprises $L$ layers, each conforming to the standard Transformer architecture as described in \cite{attentionisallyouneed}.

\color{black}
\subsection{Query Formulation}
\label{query defined}
At each agent, a set of anchor queries $\mathcal{B} \in \mathbb{R}^{M \times 8}$ ($M$ denotes the number of anchor queries) are initially randomly generated in the agent's local 3D coordinate system. Each  anchor's format is:
\[\mathcal{B}_i = \{\boldsymbol{x},\boldsymbol{y},\boldsymbol{z},\boldsymbol{h},\boldsymbol{w},\boldsymbol{l},\boldsymbol{\sin{\theta}},\boldsymbol{\cos{\theta}}\},\]
where $\boldsymbol{x}$, $\boldsymbol{y}$, and $\boldsymbol{z}$ represent the cartesian coordinates at the center of an anchor query, and $\boldsymbol{h}$, $\boldsymbol{w}$, and $\boldsymbol{l}$ specify the shape of the anchor query. The angle $\boldsymbol{\theta}$ denotes the orientation of the anchor query. Note that each anchor query can be thought as a predefined object proposal box, designed to specify potential object locations within an image. 
We use an Anchor Featuring Block (AFB) to extract image features denoted by $\mathbf{F}_i \in \mathbb{R}^{C}$ 
for the anchor query $\mathcal{B}_i$, where
$C$ denotes the dimension of the query feature.

\subsection{Anchor Featuring Block (AFB)}
In this section, we introduce how image features and anchor features interact through cross-attention to refine the anchor features.



\textbf{Local points sampling.}
Local sampling generates three kinds of points inside an anchor box $\mathcal{B}$, which are eight fixed corner points $\mathbf{p}_f$, a center point $\mathbf{p}_c$, and $\mathfrak{L}$ learnable points $\mathbf{p}_l$. $\mathbf{p}_f$ are generated by the center point $\mathbf{p}_c$ and the shape of the anchor $S$ ($\boldsymbol{w,h,l}$).
$\mathbf{p}_l$ are generated by:
\begin{equation}
    O_{f} = \mathbf{sigmoid}(\Phi_{\mathbf{linear}}(\mathbf{F})-0.5) \cdot \mathbf{R}_{yaw}
\end{equation}
\begin{equation}
    \mathbf{p}_l = O_{f} \times S + \mathbf{p}_c  \in  \mathbb{R}^{M\times \mathfrak{L}\times 3}
\end{equation}
Here $\mathfrak{L}$, and $\mathbf{R}_{yaw}$ denote the number of learnable points and orientation of anchor queries, respectively. $\mathbf{F}$ is the query features. $\Phi_{\mathbf{linear}}$ denotes a linear projection operator. We yield the point pool which contain $\mathbf{p}_c$, $\mathbf{p}_f$ and $\mathbf{p}_l$, denoted as $\mathcal{P} \in \mathbb{R}^{M\times (1+8+\mathfrak{L})\times 3}$. Intuitively,  $\mathbf{p}_c$, $\mathbf{p}_f$ and $\mathbf{p}_l$ are close in 3D space, they typically maintain this proximity when projected onto 2D space, thereby enabling the extraction of more diverse and rich features.

\textbf{Projection and Feature Integration.} We extract features $\mathbf{f}$ from multi-view images using a mainstream backbone. Then, based on the camera's projection matrix $\mathbf{P}_{\text{projection}} \in \mathbb{R}^{3 \times 4}$, the point pool is projected onto multi-view images. Specifically, the coordinates are transformed using $[u,v]^\text{T} = \mathbf{P}_{\text{projection}} \cdot [x,y,z,1]^\text{T}$. For each projection point, we utilize the Deformable-DETR \cite{deformable} approach to sample its features. Next, we implement hierarchical feature integration, using sum pooling to combine the features from multi-view points and points pool, thereby obtaining a comprehensive query feature:
\begin{equation}
    \mathbf{F} \leftarrow \sum_{i=1}^\mathcal{I}{\sum_{j=1}^{1+8+\mathfrak{L}}{\mathcal{W}_i \cdot 
    \text{MSDeformAttn}(\mathbf{f}_i,\mathcal{P})}},
\end{equation}

here, $\mathcal{I}$ denotes the number of multi-view images. 
Considering that different views have uneven importance, we incorporate encoded camera parameters (e.g., intrinsic and extrinsic) through an MLP into the feature sampling process as weights $\mathcal{W}_i$.

\subsection{Spatial-Aware Self-Attention} 
\label{SASA}
Next, the anchor features pass through the SASA module to facilitate interaction with the global information. This module incorporates spatial distance into the self-attention mechanism.
Although normal self-attention facilitates global information exchange among tokens, it proves \cite{self-attention-Computationally-expensive-1,self-attention-Computationally-expensive-2} to be an ineffective strategy in case there are a large number of anchor queries, where individual queries do not necessarily benefit from engaging with distant ones. Inspired by \cite{swintrans,sparsebev}, we introduce a simple yet effective method that considers the spatial distance between each pair of queries to dynamically select the attention map for each query. Specifically, given a set of queries $\mathcal{B}$, we  compute the Distance Matrix $D_{i,j}$ of these queries in BEV 2D space as follows:
\begin{equation}
D_{i,j} = \sqrt{(\mathbf{p}_c^i(x)-\mathbf{p}_c^j(x))^2 + (\mathbf{p}_c^i(y)-\mathbf{p}_c^j(y))^2},
\end{equation}
where $i$ and $j$ denote the indices of the queries, and $\mathbf{p}_c$ represents the center point.
We integrate $D \in \mathbb{R}^{M\times M}$ into the multi-head self-attention  to introduce a spatial-aware self-attention (SASA) module as follows:
\begin{equation}
Q = \sum_{h=1}^H \mathbf{W}_h \cdot \text{Softmax}\left(\frac{Q_h K_h^T}{\sqrt{C^\prime}} - \mathcal{D}_h\right) V_h,
\label{SASA_eq}
\end{equation}
where $h$ indexes the attention head, and $\mathbf{W}_h \in \mathbb{R}^{C\times C^\prime}$ consists of learnable weights, with $C^\prime = C/H$. $\mathcal{D}_h = \boldsymbol{\gamma}_h \cdot \log(1+D)$, where $\boldsymbol{\gamma}_h$ is a spatial-aware factor ranging from $[0,1]$, learned from the feature $\mathbf{F}$. $Q$, $K$, and $V$ represent the same query, key, and value features $\mathbf{F}$. Adopting this relatively mild adaptive distance attenuation strategy can effectively explore the spatial consistency and contextual relationships of the anchor queries.

\subsection{Anchor-based Local-Global Fusion}
Most existing collaborative perception methods directly communicate the feature maps and conduct fusion by feature maps. 
The Local-Global Fusion module leverages the spatial position information and confidence of the anchor queries to flexibly transfer anchor feature information between different agents.
Specially, we consider a collaborative scenario with $N$ agents. Given transformation matrix $\mathbf{T}_{\text{transform}} \in \mathbb{R}^{N \times N \times 3 \times 4}$, anchor queries $\mathcal{B} \in \mathbb{R}^{N \times M \times 8}$, and corresponding  query feature $\mathbf{F} \in \mathbb{R}^{N \times M \times C}$. 
We assume all $N$ agents are within the communication range. Each agent firstly evaluates a confidence score for the proposed anchors.  

\textbf{Anchor Confidence Generator.}
High-quality anchor queries provide informative clues for object detection, whereas low-quality ones can impair the original perception data. We use a detection decoder structure to produce the anchor confidence and select the top-$K$ high confidence anchors to communicate. Given the query feature $\mathbf{F}$, the corresponding anchor confidence is defined as:
\begin{equation}
\mathbf{C} = \Phi_{\text{generator}} (\mathbf{F}) \in \mathbb{R}^{N \times M \times 1}
\end{equation}
\begin{equation}
\mathbf{C}_{\text{top}}, \mathcal{B}_{\text{top}}, \mathbf{F}_{\text{top}} = \Phi_{\text{topk}}(\mathbf{C}) \in \mathbb{R}^{N \times K \times 1}
\end{equation}

where $\mathbf{C}$ represents the anchor confidence scores, measuring the possibility of containing foreground objects in each anchor. According to the index obtained above, we  select the top-$K$ anchors, denoted as $\mathcal{B}_{\text{top}} \in \mathbb{R}^{N \times K \times 8}$, along with the corresponding query feature $\mathbf{F}_{\text{top}} \in \mathbb{R}^{N \times K \times C}$ and the confidence scores $\mathbf{C}_{\text{top}}$. 
Among the top-$K$ anchors, we further filter out the anchors whose confidence scores are lower than a threshold $\tau_{\text{thre}}$ to exclude their costs and impacts in the following communication and fusion process.  
We use a binary matrix to represent whether each anchor is selected or not, where $1$ denote being selected and $0$ elsewhere:
\begin{equation}
    \mathbf{M}_{\text{confidence}} = I(\mathbf{C}_{\text{top}}, \tau_{\text{thre}})\in\{0,1\}^{N\times K \times 1}
\end{equation}
Here, $I(\cdot)$ is an indicator function. $\mathbf{M}_{\text{confidence}}$ determines which anchors will continue to participate in the fusion and which are excluded. The corresponding selected feature set is denoted: $\mathbf{F}_{\text{selected}} = \mathbf{F}_{\text{top}} \odot \mathbf{M}_{\text{confidence}}$.

\textbf{Location-aware anchor encoder.}
To facilitate fusion, it's essential to convert the anchor queries from each agent into a unified coordinate system. Given the transformation matrix $\mathbf{T}_{\text{transform}}$ and a set of anchor queries $\boldsymbol{\Omega}$ from agents other than the ego agent $i$, the received anchor queries will be projected into a unified ego cars' 3D space:
\begin{equation}
\forall j \in \Omega, \quad \mathcal{B}_{\text{selected}}^{i \leftarrow j} = \mathbf{T}_{\text{transform}}^{i \leftarrow j} \cdot \mathcal{B}_{\text{selected}}^j.
\end{equation}



Taking the spatial relationship information between agents into account, $\mathbf{T}_{\text{transform}}$ then is embeded into the query feature using a MLP layer $\Phi_{\text{MLP}}$:
\begin{equation}
    \mathbf{F}_{\text{selected}} \leftarrow \Phi_{\text{MLP}} (\mathbf{T}_{\text{transform}}) + \Phi_{\text{ae}}(\mathcal{B}_{\text{selected}}) + \mathbf{F}_{\text{selected}}.
\end{equation}
Here, $\mathbf{F}_{\text{selected}} = \{{\mathbf{F}_{\text{selected}}^{\omega}: \omega \in \Omega }\}$ represents the selected features set. $\Phi_{\text{ae}}$ encodes spatial information of the anchor into features using MLP layers. Note that $\boldsymbol{\Omega}$ does not include the ego-agent, and $\mathbf{F}_{\text{selected}}$ is defined in $\mathbb{R}^{N \times \mathcal{K} \times C}$, where $\mathcal{K} = K \times (N-1)$ represents the total number of features selected from other agents. By encoding transformation matrix $\mathbf{T}_{\text{transform}}$ into features, we can directly integrate the relative position and posture information of nearby agents into the model, allowing the model to more accurately capture the spatial relationships between agents.

\textbf{Local Anchor Alignment-based Fusion (LAAF).} For the $i$-th ego agent, given $\mathbf{F}_i$ and $\mathcal{B}_i$, and the sets $\mathbf{F}_{\text{selected}}$ and $\mathcal{B}_{\text{selected}}$ from neighboring agents, local fusion is conducted by aligning the anchor queries and aggregating the anchor features. This is accomplished using $\mathcal{M}$, a non-parametric deterministic mapping function that produces two range points to represent the extent of the fusion field: $\mathbf{p}_{\min,m}, \mathbf{p}_{\max,m} = \mathcal{M}(\mathcal{B}_{i,m})$. 
Here, the range points $\mathbf{p}_{\min,m}$ and $\mathbf{p}_{\max,m}$ represent the vertices at the opposite ends of the diagonal spanning a rectangular prism for the $m$-th anchor of $\mathcal{B}_i$. Then, for each anchor feature $\mathbf{F}_{\text{select},j}$ in $\Omega$, whose center is contained within the receptive field, we simply sum the anchor features within the receptive field:
\begin{equation}
    \mathbf{F}_{i,m} \leftarrow \mathbf{F}_{i,m} + \sum_{j \in J_m} \mathbf{F}_{\text{selected}, j},
\end{equation}
where $J_m=\{j:\mathbf{p}_{\text{selected},j}\subseteq[\mathbf{p}_{\min,m},\mathbf{p}_{\max,m}]\}$. $\mathbf{p}_{\mathrm{selected}}$ is the center point from $\mathcal{B}_{\text{selected}}$ and $J_m$ is the set of indices $j$ for which $\mathbf{p}_{\text{selected},j}$ falls within the $m$-th receptive field defined by $\mathbf{p}_{\min,m}$ and $\mathbf{p}_{\max,m}$. 

\textbf{Spatial-Aware Cross-Attention (SACA).} 
The local anchor-level fusion integrates local information, while SACA facilitates interactions between the ego anchor queries and those of all neighboring agents, achieving global information enhancement. Similar to SASA \ref{SASA}, we adopt the same strategy  by calculating the distance between the selected anchor and the ego anchor. 
Additionally, here, the query is the ego anchor feature, and the key is $\mathbf{F}_{\text{selected}}$. $\mathcal{D}_h$ here represents the spatial difference matrix between the anchor $\mathcal{B}$ and $\mathcal{B}_{\text{select}}$, obtained using the same formula as in SASA Eq. \ref{SASA_eq}.
Linux
After local-global fusion, the fused features will enter a decoder to adjust the anchor proposals. Above process will repeat $L$ layers to output the final predictions. 

\color{black}

\subsection{Detector and losses.}
Finally, for the $l$-th layer, the $i$-th agent and the $m$-th anchor query $\mathcal{B}_{i,m}^l \in \mathbb{R}^{N \times M  \times 8}$ and $\mathbf{F}_{i,m}^l$, we predict a bounding box $\hat{\mathbf{b}}^l_{i,m}$, its categorical label $\hat{\mathbf{c}}_{i,m}^l$ with two neural networks $\Phi_{\text{reg}}^l$ and $\Phi_{\text{cls}}^l$:
\begin{equation}
    \hat{\mathbf{b}}^l_{i,m} = \Phi_{\text{reg}}^l(\mathcal{B}_{i,m}^l), \qquad
    \hat{\mathbf{c}}_{i,m}^l = \Phi_{\text{cls}}^l(\mathbf{F}_{i,m}^l)
\end{equation}
During inference, we only use the outputs from the last layer. Following \cite{detr,Detr3D}, we use a set-to-set loss to measure the discrepancy between the prediction set $\hat{\mathcal{O}}^l (\hat{\mathbf{b}}^l_{i,m},\hat{\mathbf{c}}_{i,m}^l)$ and the ground-truth set ${\mathcal{O}}$. The ground-truth are assigned to anchor queries based on the Hungarian matching\cite{hungarian}. Mathematically, for the $l$-th decoder layer $(l = 1, ..., L)$, this process is formulated as:
\begin{equation}
    \hat{\sigma}^{(l)}=\underset{\sigma^{(l)}\in\mathfrak{S}^{(l)}}{\operatorname*{\arg\min}}\sum_{j=1}^M\mathcal{L}\left(\hat{\mathcal{O}}_j^l,{\mathcal{O}}_{\sigma^{(l)}(j)}\right),
\end{equation}
where $\sigma^{(l)}$ denotes a sampled matching combination. $\mathfrak{S}$ denotes the matching space containing all possible matching combinations between the predictions and the ground truth. $\mathcal{L}$ is matching cost, and $M$ is the number of anchor queries. $\hat{\sigma}^{(l)}$ represents the obtained optimal matching result. Then the loss for 3D object detection can be summarized as:
\begin{equation}
    L = \sum_{l=1}^L(\lambda_{cls} \cdot  \mathcal{L}_{\text{cls}}(c,\hat{c}(\hat{\sigma}^{(l)})) + \lambda_{reg} \cdot \mathcal{L}_{\text{reg}}(b,\hat{b}(\hat{\sigma}^{(l)}))).
\end{equation}
$\mathcal{L}_{\text{cls}}$ denotes the focal loss for classification, while $\mathcal{L}_{\text{reg}}$ represents the $L1$ loss for regression. The parameters $\lambda_{\text{cls}}$ and $\lambda_{\text{reg}}$ serve as different weights to balance these losses.

\begin{figure*}
    \centering
    \includegraphics[width=0.9\linewidth]{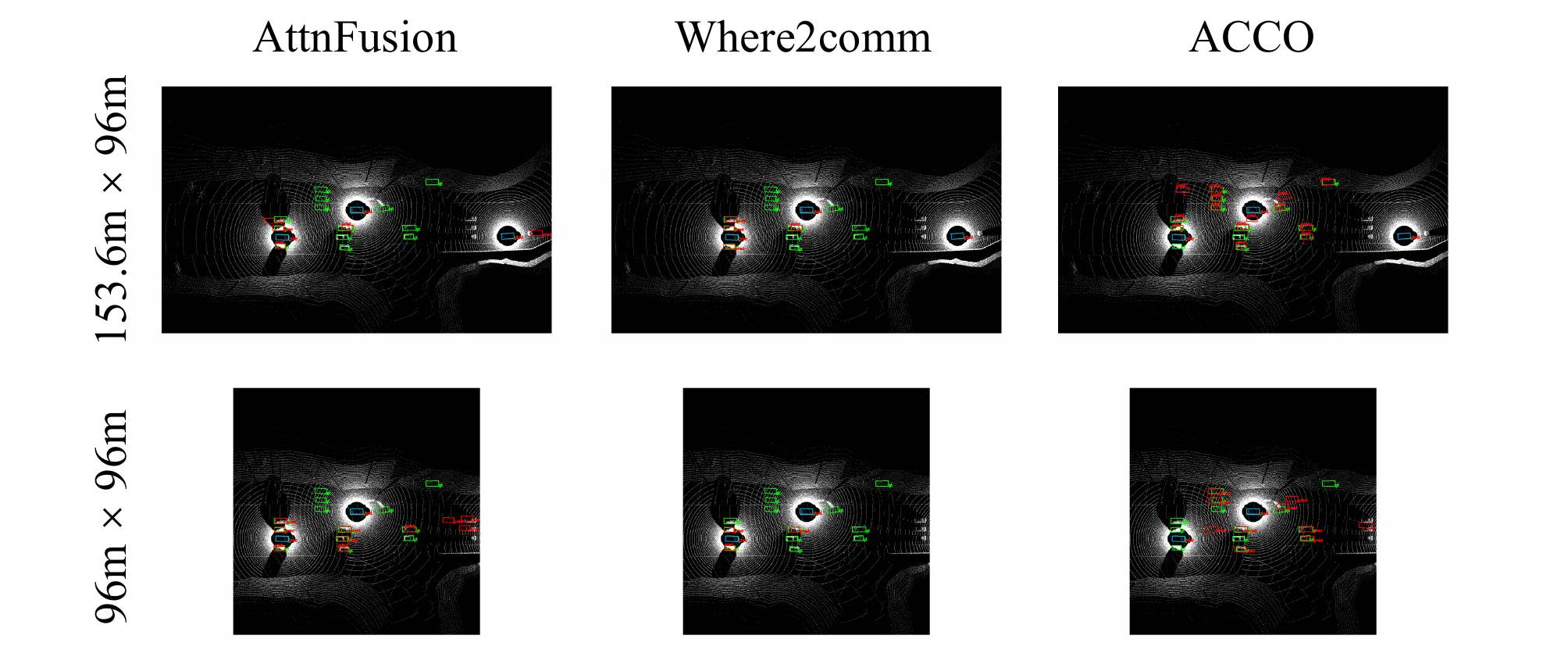}
    \caption{This visualization compares different methods applied to the OPV2V dataset. 
{\color{green}Green} and {\color{red}red} 3D bounding boxes represent the groun truth and prediction respectively. \textcolor[rgb]{0,0.749,1}{Blue} 3D bounding boxes represent the communication agents.}
    \label{visualize for res}
\end{figure*}

\section{EXPERIMENT}
\label{exp}
To thoroughly evaluate ACCO, we selected two mainstream datasets: the real-world dataset Dair-V2X \cite{dair-v2x} and the simulated dataset OPV2V \cite{OPV2V}. Performances are evaluated using Average Precision (AP) metrics at Intersection-over-Union (IoU) thresholds of $0.30$, $0.50$, and $0.70$.
\subsection{Experimental settings}
\label{expsetting}
The backbone of ACCO and anchor queries settings follow the settings described in Sparse4D \cite{sparse4D}. We employ ResNet50 \cite{ResNet} and FPN \cite{FPN} to extract image features, and use uniform initialization to set the initial $\boldsymbol{x}$, $\boldsymbol{y}$ and $\boldsymbol{z}$ coordinates of the anchors. The remaining attributes of the anchor query are set to $\{1,1,1,1,0\}$. The query feature $\mathbf{F}$ is initialized to all zeros. $\tau_{\text{thre}}$ is set to $0.5$. The ACCO encoder contains 6 encoder layers and constantly refines the anchor queries in each layer. By default, we train our models with 30 epochs, using learning rate $6\times10^{-5}$. Our method is implemented in PyTorch \cite{pytorch}. The network is trained on a NVIDIA RTX 3090 GPU(24G).

\begin{figure}[!tbp]
\centering
  \includegraphics[width=0.7\linewidth]{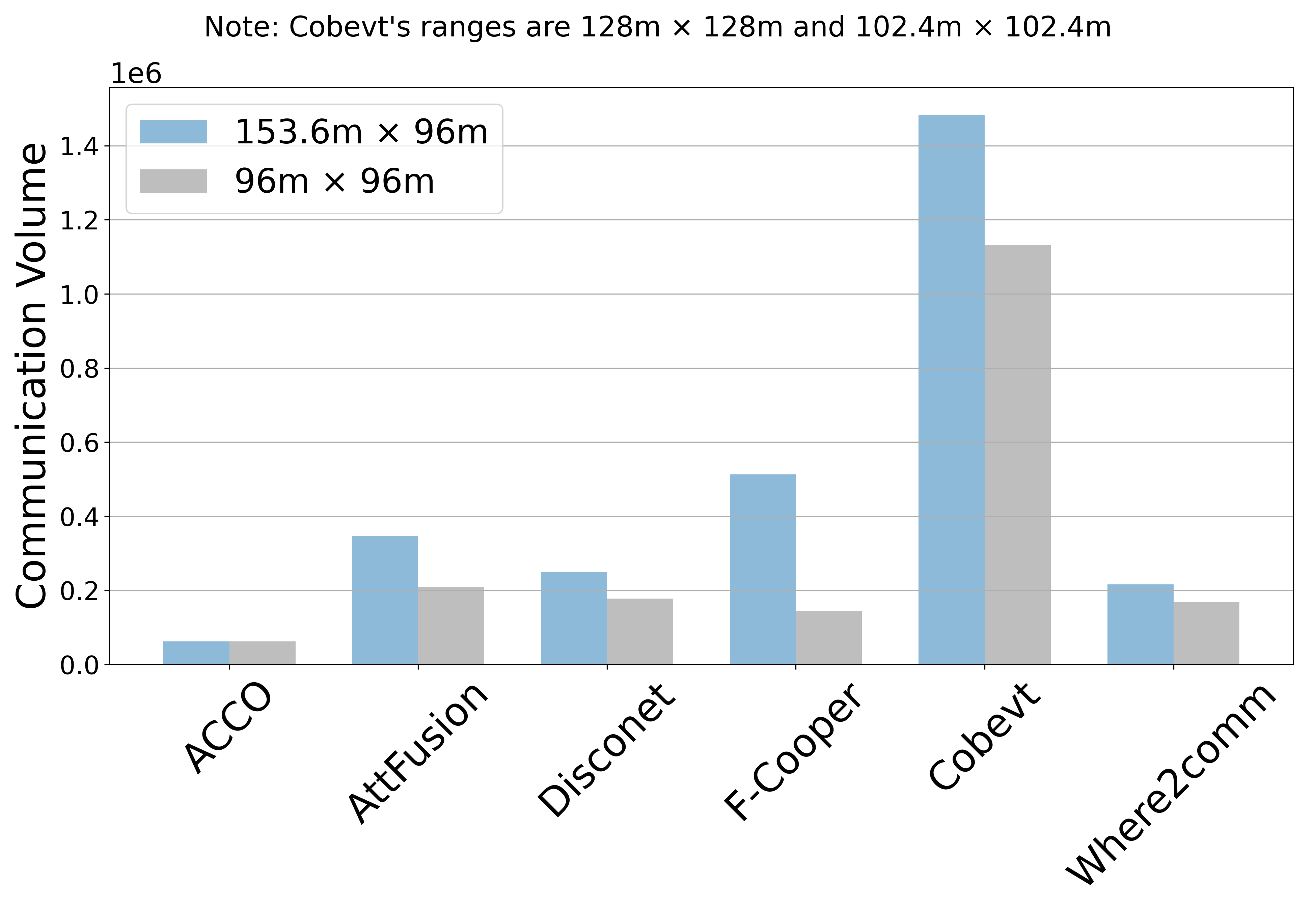}
  
  \vspace{0mm}
  \caption{Analysis of communication bandwidth across different perception distances.}
  \label{bandwidth}
\end{figure}

\begin{figure}[!htb]
    \centering
    \begin{minipage}[t]{0.49\linewidth}
        \centering
        \includegraphics[width=\linewidth]{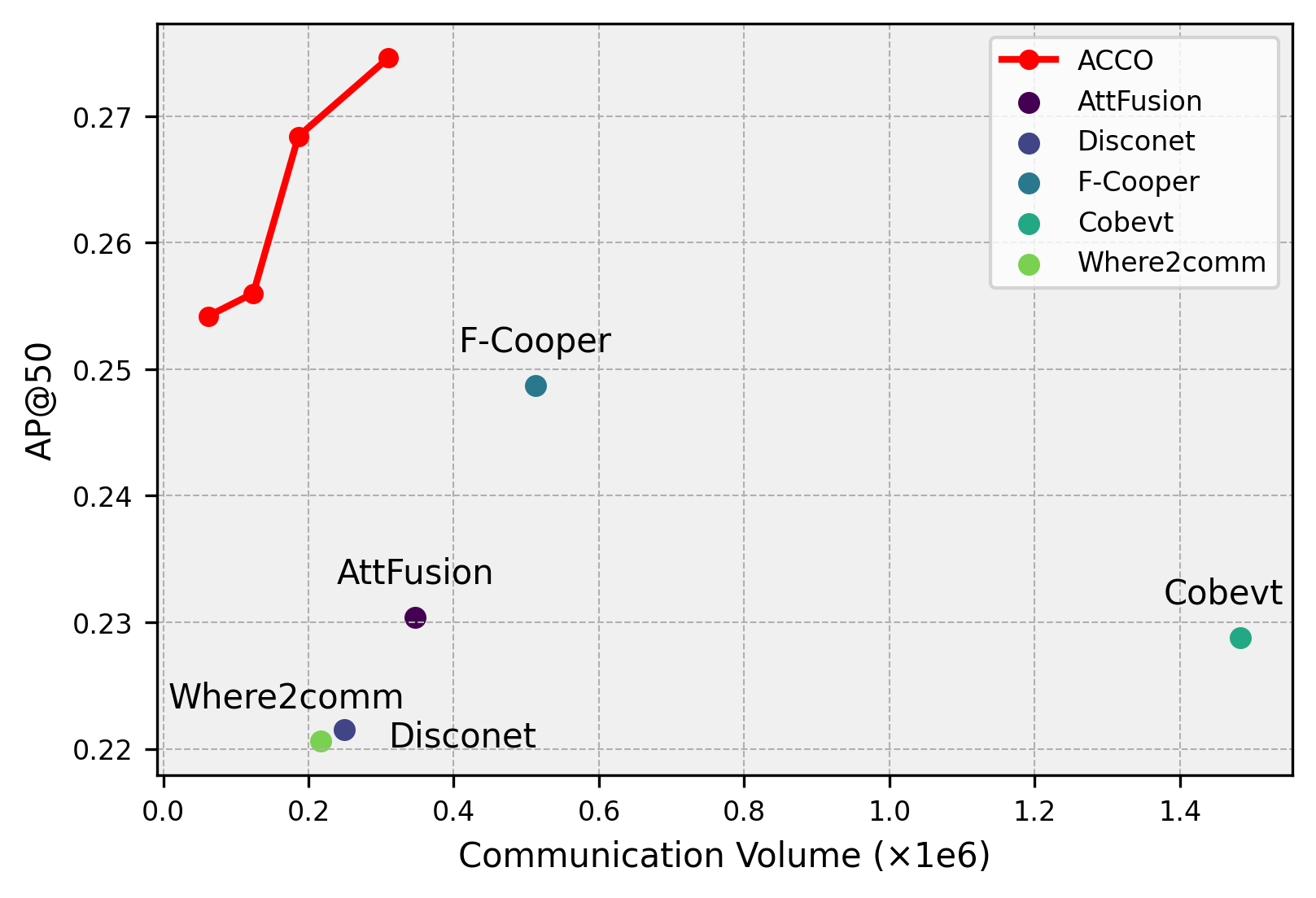} 
        \label{fig:sub1}
    \end{minipage}
    \hfill
    \begin{minipage}[t]{0.49\linewidth}
        \centering
        \includegraphics[width=\linewidth]{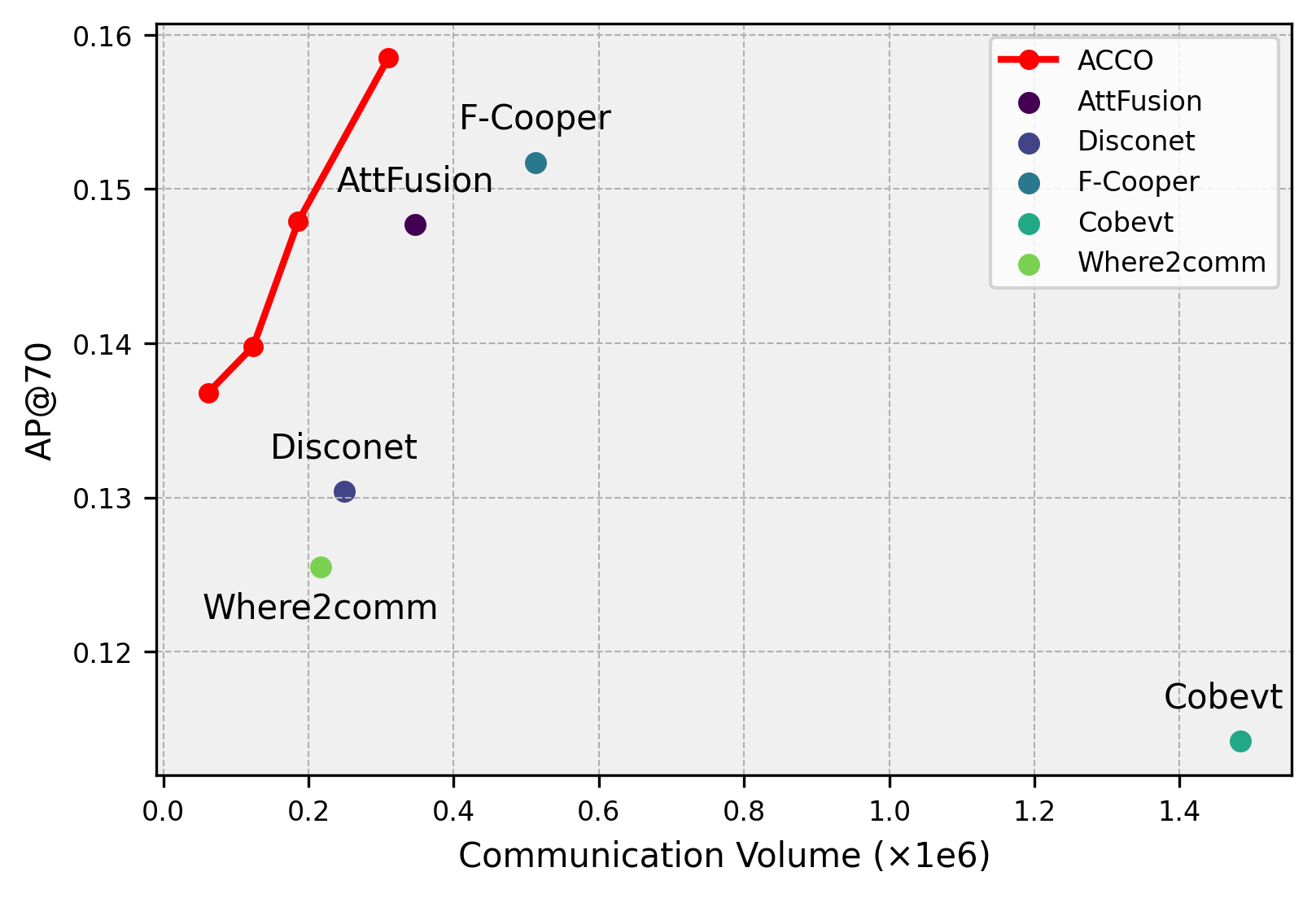} 
        \label{fig:sub2}
    \end{minipage}
    \vspace{-5mm}
    \caption{The relationship between communication volume and performance demonstrated.}
    \label{fig:test}
\end{figure}

\subsection{Quantitative evaluation}
\textbf{Evaluation on the benchmark.} Tab \ref{OPV2V_compare} and Tab \ref{dair-v2x_res} compares the proposed method with previous methods, focusing on the trade-off between detection performance (AP@IoU) and detection range. Detection range, in this context, refers to a matrix range centered on the ego vehicle in the BEV perspective. 
We compare the results with Attfusion \cite{OPV2V}, Disconet \cite{disconet}, F-Cooper \cite{Fcooper}, Cobevt \cite{CoBEVT}, and Where2comm \cite{where2comm}. We implement these methods based on LSS \cite{LSS}. The feature map is transformed to BEV with the resolution of 0.4m/pixel.
Tab.\ref{OPV2V_compare} demonstrates that, compared to other methods, our approach minimally loses accuracy across different perception ranges and consistently performs well, even at extended detection ranges. At a perception range of $153.6\text{m} \times 96\text{m}$, while traditional methods show a sharp decline in performance, our method continues to exhibit high levels of performance. Tab \ref{dair-v2x_res} shows that in the real world, under noisy conditions, our method surpasses the state-of-the-art (SOTA) performance by 12.22$\%$, 36.22$\%$, and 9.87$\%$ on AP@30, 50, and 70, respectively.

\textbf{Communication bandwidth analysis.} Fig \ref{bandwidth} shows a performance analysis of the communication volume. The results indicate that 1) methods based on feature maps inherently need avoid high communication volume, and 2) at larger detection ranges, feature map methods increase dramatically the communication volume, whereas ACCO's communication volume does not significantly increase. Fig. \ref{fig:test} illustrates the performance achieved through our multi-layer iterative communication. The fusion module can be added at any layer within the L layers. The detection performance reaches its peak when the fusion module is incorporated into layers 1 through 5.


\begin{table}[htbp]
\centering
\caption{Comparison of different methods in OPV2V dataset. \textbf{Bold} values indicate the best performance. {\color[HTML]{3166FF}Blue} and {\color[HTML]{FE0000}red} values indicate the second-best performance in different detection ranges, respectively.}
\label{OPV2V_compare}
\begin{minipage}{\linewidth}

\setlength{\tabcolsep}{6pt} 
\renewcommand{\arraystretch}{1.2} 
\small  

\resizebox{\linewidth}{!}{
\begin{tabular}{l|ccc|c}
\toprule
Method                            & \textbf{AP@30} & \textbf{AP@50} & \textbf{AP@70} & Detection range         \\ 
\midrule
\multirow{2}{*}{No collaboration} & 0.2941      & 0.2197      &   0.1099  & 153.6m $\times$ 96m      \\
                                  & 0.2631      &  0.1859     &  0.1145     & 96m $\times$ 96m    \\ 
\hline
\multirow{2}{*}{AttFusion \cite{OPV2V}}        & 0.2829 & 0.2304 & 0.1477 & 153.6m $\times$ 96m     \\ 
                                  & 0.3669 & \textbf{0.2967} & \textbf{0.1773} & 96m $\times$ 96m      \\   \cline{1-4} 

\multirow{2}{*}{Disconet \cite{disconet}}         & 0.2905 & 0.2215 & 0.1304 & 153.6m $\times$ 96m     \\    
                                  & {\color[HTML]{FE0000}0.3639} & 0.2915 & 0.1605 & 96m $\times$ 96m     \\  \cline{1-4} 

\multirow{2}{*}{F-Cooper \cite{Fcooper}}         & 0.2824 & {\color[HTML]{3166FF}0.2487} & {\color[HTML]{3166FF}0.1517} & 153.6m $\times$ 96m     \\  
                                  & 0.3623 & {\color[HTML]{FE0000}0.2931} & {\color[HTML]{FE0000}0.1753} & 96m $\times$ 96m      \\   \cline{1-4} 

\multirow{2}{*}{Cobevt \cite{CoBEVT}}           & 0.2659 & 0.2288 & 0.1142 &  128m $\times$ 128m      \\    
                                  &  0.3461     &   0.2748    &  0.1440      & 102.4m $\times$ 102.4m    \\ \cline{1-4} 

\multirow{2}{*}{Where2comm \cite{where2comm}}          & {\color[HTML]{3166FF}0.3014 }     & 0.2206      &  0.1255     & 153.6m $\times$ 96m      \\ 
                                  &  0.2795     &   0.2045    &  0.1395      & 96m $\times$ 96m      \\ 
\hline

\multirow{2}{*}{ACCO}  & \textbf{0.3689}      & \textbf{0.2746}      &   \textbf{0.1585}         & 153.6m $\times$ 96m                   \\ 

 & \textbf{0.3763}                       &  0.2873                      &   0.1652                      & 96m $\times$ 96m          \\

\bottomrule
\end{tabular}
}
\end{minipage}
\begin{minipage}{\linewidth}
\footnotesize
\vspace{1mm}
$^{*}$The detection ranges for \textit{Cobevt} are different due to model limitation.
\end{minipage}
\end{table}

\vspace{-3mm}

\subsection{Visualization}
As perception distance increases, the uncertainty in the depth distribution interval grows, which undermines the quality of traditional BEV features. Fig. \ref{visualize for res} shows the visualization results compared to \textit{AttFusion} and \textit{Where2comm}. ACCO offers distinct advantages in long-distance perception, effectively utilizing perspective information from multiple agents to address challenges such as occlusions and distance, which are difficult for conventional methods.

\begin{table}[htbp]
\centering
\caption{Comparison of different methods in Dair-v2x dataset.}
\label{dair-v2x_res}
\begin{minipage}{\linewidth}
\renewcommand{\arraystretch}{1.2} 
\setlength{\tabcolsep}{6pt} 

\resizebox{\linewidth}{!}{%
\begin{tabular}{l|ccc|c}
\toprule
\textbf{Method}           & \textbf{AP@30} & \textbf{AP@50} & \textbf{AP@70} & Detection range \\ \midrule
No collaboration &   0.0898    &  0.0781     & {\color[HTML]{3166FF}0.0729}      & \multirow{8}{*}{\rule{0pt}{3ex} 204.8m $\times$ 102.4m}  \\ \cline{1-4}  
AttFusion        &  0.0884     &  0.0719     &  0.0671     &                                                 \\ \cline{1-4}
Disconet         &  0.0749     &  0.0664     &  0.0614     &                                                 \\ \cline{1-4}
F-Cooper         &  0.0876     &  0.0672     &  0.0617     &                                                 \\ \cline{1-4}
Cobevt           &  0.0778     &  0.0675     &   0.0573    &                                                 \\ \cline{1-4}
Where2comm       & {\color[HTML]{3166FF}0.1006}      & {\color[HTML]{3166FF}0.0784}      &  0.0656     &                                                 \\ \cline{1-4}  
ACCO             & \textbf{0.1129}      & \textbf{0.1068}      & \textbf{0.0801}      &                                                \\

\bottomrule
\end{tabular}
}
\end{minipage}
\vspace{-5mm} 
\end{table}

\subsection{Ablation}

\textbf{Ablation studies on hyperparameters.} Tab \ref{number of anchor} presents an ablation study on the number of anchor queries. The experiments indicate that performance stabilizes beyond 600 anchor queries. Considering the balance between performance, computational efficiency, and memory usage, we selected 600 as the optimal baseline setting. Tab \ref{Topk} illustrates the selection of Top-$K$ anchors. We can observe that: 1) Smaller $K$ fails to fully utilize the information from neighboring agents, resulting in suboptimal performance. 2) Larger $K$ may transmit noisy information, hindering feature fusion. The experiments indicate that the best performance is achieved when $K$ is set to 10. Tab \ref{number of agents} shows the impact of the number of agents. Clearly, more agents provide a richer perspective, thereby improving detection performance.

\begin{table}[!htbp]
\centering
\caption{Ablation studies.}
\label{tab:combined-ablation}
\scriptsize 

\begin{subtable}{0.24\textwidth} 
\centering
\caption{Number of anchor queries ($M$)}
\begin{tabular}{l|ll}
\toprule
$M$ & \textbf{AP@50} & \textbf{AP@70} \\ \hline
300        &   0.2703             &   0.1492             \\
600        &   \textbf{0.2746}           &  0.1585                \\
900        &   0.2734         &    \textbf{0.1601}           \\ \bottomrule
\end{tabular}
\label{number of anchor}
\end{subtable}
\hfill 
\begin{subtable}{0.24\textwidth} 
\centering
\caption{Top-$K$ anchor selected ($K$)}
\begin{tabular}{l|ll}
\toprule
$K$ & \textbf{AP@50} & \textbf{AP@70} \\ \hline
5          &  0.2462              &   0.1250             \\
10          &     \textbf{0.2746}           &  \textbf{0.1585}                 \\
15         &      0.2592           &   0.1370             \\
\bottomrule
\end{tabular}
\label{Topk}
\end{subtable}
\hfill 
\begin{subtable}{0.24\textwidth} 
\centering
\caption{Number of agents ($N$)}
\begin{tabular}{l|ll}
\toprule
$N$ & \textbf{AP@50} & \textbf{AP@70} \\ \hline
2          &    0.2437            &  0.1082              \\
3          &    0.2384            &  0.1120              \\
4          &    0.2574            &   0.1342             \\
5          &     \textbf{0.2746}           &  \textbf{0.1585}                \\ \bottomrule
\end{tabular}
\label{number of agents}
\end{subtable}
\hfill 
\begin{subtable}{0.24\textwidth} 
\centering
\caption{Components Analysis}
\begin{tabular}{@{}ccc|c@{}}
\toprule
ACG & LAAF & SACA & OPV2V \\ \hline   
\ding{55} & \ding{55} & \ding{55} & 21.97/10.99 \\
\ding{55} & \checkmark & \ding{55} & 24.87/12.53 \\
\checkmark & \checkmark & \ding{55} & 27.43/15.24 \\
\checkmark & \checkmark & \checkmark & \textbf{27.46/15.85} \\ \bottomrule
\end{tabular}
\label{component}
\end{subtable}
\end{table}

\textbf{Component analysis.} 
Tab \ref{component} evaluates several key components. We can see that: 1) LAAF significantly enhances the performance. This means that it can effectively integrate anchor-level features to address issues of occlusion and long distances. 2) Compared to normal attention mechanisms, SACA incorporates spatial information, which enhances performance. 3) ACG enhances performance by providing more high-quality anchor queries. The three designed components resulted in performance improvements of 31.23\%, 27.54\%, and 43.40\% on AP@30, 50, and 70, respectively.

\section{Conclusion}
\label{Conculusion and limitation}
In this paper, we present ACCO, a novel camera-based fusion strategy that addresses accuracy and detection range challenges in multi-agent collaborative perception. By using a transformer architecture for anchor query-based fusion, ACCO enhances perception, reduces communication bandwidth, and improves performance. Experimental results demonstrate a better balance between bandwidth and performance, as well as an expanded detection range. Future work will apply this method to more collaborative perception datasets and other sensor modalities.

\section*{ACKNOWLEDGMENT}
This work was supported by the National Natural Science Foundation of China Grant No. 12071478, No. 61972404; Public Computing Cloud and the Blockchain Lab, School of Information, Renmin University of China.

\clearpage
{
    \small
    \bibliographystyle{IEEEtran}
    \bibliography{main}
}

\end{document}